\documentclass[10pt]{article}

\usepackage{wsstyle}

\usepackage[T1]{fontenc}
\usepackage{mathptmx}
\usepackage[scaled=0.92]{helvet}
\usepackage{courier}
\usepackage{microtype}
\usepackage{amsmath,amssymb}
\usepackage{array}
\usepackage{booktabs}
\usepackage{graphicx}
\usepackage[numbers,sort&compress]{natbib}
\usepackage[colorlinks=true,allcolors=blue,breaklinks=true]{hyperref}
\usepackage{url}

\usepackage{etoolbox}
\apptocmd{\thebibliography}{%
  \raggedright
}{}{}

\makeatletter
\AtBeginDocument{%
  \def\do@url@hyp{}%
  \Urlmuskip=0mu\relax
}
\makeatother

\newcommand{\NRuns}{50}
\newcommand{\SOneAtoaLocalMed}{2.8}

\newcommand{\SOneAtoaLocalBytesUp}{170}
\newcommand{\SOneAtoaLocalBytesDown}{569}

\newcommand{\SOneAtoaRttMed}{61.0}

\newcommand{\SOneAcpLocalMed}{2.8}

\newcommand{\SOneAcpRttMed}{62.1}

\newcommand{\SOneMcpLegacyLocalMed}{6.1}

\newcommand{\SOneMcpLegacyLocalConns}{3}
\newcommand{\SOneMcpLegacyLocalBytesUp}{1147}
\newcommand{\SOneMcpLegacyLocalBytesDown}{1522}

\newcommand{\SOneMcpLegacyRttMed}{230}
\newcommand{\SOneMcpLegacyRttPNF}{236}

\newcommand{\SOneMcpModernLocalMed}{4.4}

\newcommand{\SOneMcpModernRttMed}{122}

\newcommand{\STwoAtoaWarmRttMed}{2.0}

\newcommand{\STwoAcpWarmRttMed}{6.6}

\newcommand{\STwoMcpLegacyWarmRttMed}{230}

\newcommand{\STwoMcpLegacyWarmRttRts}{3}

\newcommand{\STwoMcpModernWarmRttMed}{63.5}

\newcommand{\STwoMcpModernWarmRttRts}{1}

\newcommand{\STwoMcpPinnedWarmRttMed}{6.6}

\newcommand{\SThreeAtoaRttMed}{62.2}

\newcommand{\SThreeAcpRttMed}{57.7}

\newcommand{\SThreeMcpLegacyRttMed}{231}

\newcommand{\SThreeMcpModernRttMed}{117}

\newcommand{\SFourAAtoaLocalMed}{2011}

\newcommand{\SFourAAcpLocalMed}{2011}

\newcommand{\SFourAMcpLegacyLocalMed}{6026}

\newcommand{\SFourAMcpModernLocalMed}{4019}

\newcommand{\SFourCAtoaLocalMed}{3.0}

\newcommand{\SFourCMcpLegacyLocalTtfMed}{3.7}

\newcommand{\SFourCMcpModernLocalTtfMed}{4.2}

\newcommand{\TokMcpLegacy}{181}

\newcommand{\TokMcpModern}{242}

\newcommand{\TokAcp}{112}

\newcommand{\SlopeMcpLegacy}{4.2}
\newcommand{\SlopeMcpModern}{2.1}
\newcommand{\SlopeAtoa}{1.1}
\newcommand{\SlopeAcp}{1.1}
\newcommand{\SweepN}{20}
\newcommand{\SweepVsTableMaxDeltaMs}{2.4}
\newcommand{\CiMaxHalfWidth}{2.0}
\newcommand{\CiMedHalfWidth}{0.16}
\newcommand{\CiIters}{2{,}000}
\newcommand{\InvPerCapMcp}{78}

\newcommand{\InvMcpNTwoHundred}{15{,}618}
\newcommand{\InvPerCapAtoa}{24}

\newcommand{\InvPerCapAcp}{107}

\newcommand{\InvAcpNTwoHundred}{21{,}405}
\newcommand{\InvMaxCtxPct}{11}
\newcommand{\InvMaxN}{200}
\newcommand{\BaselineTcpMed}{0.05}
\newcommand{\BaselineTlsMed}{1.88}
\newcommand{\HeadlineColdRatio}{3.8}
\newcommand{\McpModernSpeedup}{47}
\newcommand{\AmortMcpLegacy}{230}
\newcommand{\AmortMcpModern}{64.1}
\newcommand{\AmortMcpPinned}{7.8}
\newcommand{\AmortAtoa}{2.6}
\newcommand{\AmortAcp}{7.1}

\newcommand{\mcpl}{MCP (legacy)}
\newcommand{\mcpm}{MCP (modern)}

\newcommand{\code}[1]{\mbox{\texttt{#1}}}
\newcommand{\pth}[1]{\path{#1}}

\title{The Cost of Saying Hello:\\[0.25em]{\large An Empirical Study of
Handshake Overhead in AI Agent Communication Protocols}}

\author{%
  Pranav Varshney\\
  University of Michigan\\
  \texttt{pvarsh@umich.edu}%
}
\date{}

\begin{document}

\maketitle

\begin{abstract}
\noindent
Every interaction between AI agents begins with a handshake. The
initiating agent must locate its counterpart, learn what it can do, and
establish a session before the first useful message can be sent. Surveys
of agent communication protocols name handshake overhead as an efficiency
criterion, yet none report measurements. This paper presents the first
empirical, phase-decomposed benchmark of the handshake
in three major agent protocols: MCP (in both its 2025 \emph{initialize}
and 2026 \emph{discover} eras), A2A, and ACP. Using each protocol's
official SDK, we instrument every request for round trips, per-phase
latency, wire bytes, and token cost across cold-start, warm-repeat,
capability-mismatch, and degraded-metadata scenarios ($N{=}\NRuns{}$ per
cell, on loopback and under an emulated 50\,ms RTT). Round-trip count
dominates handshake cost. A legacy MCP handshake takes three round trips
and \SOneMcpLegacyRttMed{}\,ms at 50\,ms RTT; A2A needs one round trip
and \SOneAtoaRttMed{}\,ms. Round trips are not the whole bill: when an
LLM reads the counterpart's metadata, handshake cost scales with the
counterpart's capability inventory at \InvPerCapAtoa{}--\InvPerCapAcp{}
tokens per capability, so a \InvMaxN{}-capability counterpart spends
\InvMaxCtxPct{}\% of a 200k-token context window before the agent reads
its own task. MCP's own 2026
revision cuts its handshake from three round trips to two and, with era
pinning, to zero on warm reconnects, protocol evolution that corroborates
the cost we measure. We release the full harness, raw data, and analysis
at \url{https://github.com/pvarshh/handshake-benchmark}.
\end{abstract}


\section{Introduction}
\label{sec:intro}

Software agents built on large language models are beginning to interact
with other agents as a matter of course. A coding assistant calls a tool
server; an orchestrator hands a subtask to a specialist; a consumer agent
queries a merchant agent. Each of these interactions is preceded by a
connection-establishment exchange in which the initiating agent discovers
its counterpart, learns its capabilities, and establishes a session. We
call this exchange the \emph{handshake}, by analogy with transport-layer
connection setup. The handshake is pure overhead. No useful work happens
until it completes, and in open agent ecosystems, where counterparts are
strangers, capabilities are heterogeneous, and interactions are often
short, its cost is paid frequently and amortized poorly.

The systems community has a long tradition of measuring exactly this kind
of tax at lower layers. The TCP three-way handshake, the TLS key
exchange~\citep{rescorla2018tls,lee2021tls,naylor2014cost}, and QUIC's
0-RTT resumption~\citep{iyengar2021quic} were all designed, measured, and
redesigned around one insight: connection establishment is a first-order
cost of short interactions. The emerging stack of AI agent communication
protocols has not yet received the same treatment. The most comprehensive
survey of the space names ``handshake overhead'' as an explicit
efficiency criterion for comparing protocols, yet reports no
measurements~\citep{yang2025survey}. Recent benchmarks measure agent
protocols end-to-end at the task level~\citep{du2026protocolbench,
dobrovolskyi2026empirical}, but none isolates the
connection-establishment phase itself.

This paper fills that gap. We benchmark the handshake, and only the
handshake, of three major agent communication protocols head-to-head,
using their official SDKs: the Model Context Protocol
(MCP)~\citep{anthropic2024mcp,mcpspec2026}, the Agent2Agent protocol
(A2A)~\citep{a2aspec2026}, and the Agent Communication Protocol
(ACP)~\citep{acpspec2025}. ACP merged into A2A in August 2025 and is
frozen; we include it because its deployments persist and because its
one-round-trip REST design is a distinct point in the handshake design
space, which a comparison of only the two live protocols would miss. For
MCP we measure two protocol eras
separately. The \emph{legacy} handshake (specification revisions 2024-11
through 2025-11) is what nearly all deployed MCP software used at the
time of writing; the \emph{modern} handshake, introduced in revision
2026-07-28, replaces the three-message initialization with a single
cacheable probe. This within-protocol comparison turns a recent design
change into a natural experiment on the cost of handshake round trips.

Our contributions are:
(C1)~a formal decomposition of the agent handshake into discovery,
capability-exchange, session, and readiness phases, with a precise
per-protocol definition of the \emph{task-ready} state
(Section~\ref{sec:defs});
(C2)~to our knowledge the first cross-protocol empirical benchmark of
handshake overhead, covering round trips, phase-decomposed latency, wire
bytes, capability-metadata token cost, and failure behavior across four
scenario families and two network conditions, with $N{=}\NRuns$ runs per
cell (Sections~\ref{sec:method}--\ref{sec:results});
(C3)~an open reproducibility package (harness, minimal counterpart
agents, raw per-run logs, and analysis scripts) with one-command
reproduction; and
(C4)~design implications for protocol and agent-framework authors,
grounded in where the measured overhead actually concentrates
(Section~\ref{sec:discussion}).

Our headline findings are simple. First, the moment a real network is
involved, round-trip count dominates handshake cost; payload size and
compute do not. Under a 50\,ms RTT, the three-round-trip legacy MCP
handshake costs \SOneMcpLegacyRttMed{}\,ms to task-ready (median), while
A2A's single card fetch costs \SOneAtoaRttMed{}\,ms, a factor of
\HeadlineColdRatio{}. Second, protocols differ sharply in what a warm
reconnection may skip. A2A, ACP, and modern MCP can reach task-ready with
zero round trips using spec-sanctioned caching (modern MCP when pinned to
a cached protocol era; its default policy still revalidates in one),
while legacy MCP repeats its full handshake on every connection. Third,
failure behavior diverges more than success behavior. All three protocols
detect malformed metadata quickly, but only MCP validates version claims
during the handshake itself. A2A, by design, defers version checking to a
per-request header; its client completed tasks against a counterpart
whose card claims protocol version 99.0.0, and a real incompatibility
would surface only at the first task message.

\section{Background and Related Work}
\label{sec:background}

\subsection{The protocols under study}

\textbf{MCP.} The Model Context Protocol, introduced by Anthropic in
November 2024~\citep{anthropic2024mcp} and now governed as a Linux
Foundation project, connects a client (typically an LLM application) to
servers exposing tools, resources, and prompts. It is a JSON-RPC
protocol; we use its streamable HTTP transport. Through specification
revision 2025-11-25 the handshake comprises an \code{initialize}
request/response carrying protocol version and capability flags, an
\code{initialized} notification, and, for a client that needs the tool
inventory, a \code{tools/list} request~\citep{mcpspec2025}. Revision
2026-07-28 introduces a second era. A single \code{server/discover}
probe returns versions, capabilities, and instructions in one exchange,
and responses to discovery and listing requests carry explicit freshness
hints (\code{ttlMs}, \code{cacheScope}; SEP-2549) that make them
client-cacheable across sessions~\citep{mcpspec2026}. We measure both
eras.

\textbf{A2A.} The Agent2Agent protocol, announced by Google in April 2025
and donated to the Linux Foundation, targets peer-to-peer task delegation
between opaque agents. Discovery is document-centric: an agent publishes
an \emph{Agent Card} at a well-known URI
(\pth{/.well-known/agent-card.json}) describing its identity,
supported interfaces, and skills, and task exchange then proceeds over a
JSON-RPC, REST, or gRPC binding~\citep{a2aspec2026}. We benchmark
specification v1.0.0 with the JSON-RPC binding.

\textbf{ACP.} The Agent Communication Protocol, from IBM's BeeAI project,
is REST-first. A server exposes \texttt{GET /agents}, returning manifests
for the agents it hosts, and runs are created with \texttt{POST /runs}.
In August 2025 the ACP project merged into A2A under the Linux Foundation
and its repository was archived~\citep{lfaidata2025acp}; the
specification and SDK remain published and deployed~\citep{acpspec2025}.
We include ACP because deployments persist and because its one-round-trip
REST design occupies a distinct point in the handshake design space. We
treat it throughout as a frozen protocol.

\subsection{Related work}

Surveys of the agent-protocol landscape compare these protocols
qualitatively. \citet{yang2025survey} propose evaluation dimensions,
efficiency and handshake overhead explicitly among them, but decline to
define a benchmark. \citet{ehtesham2025survey} compare MCP, ACP, A2A, and
ANP architecturally. Security analyses have compared the protocols'
threat surfaces~\citep{louck2025security,anbiaee2026threat,hou2025mcp}.
Performance has received far less attention.

Two recent empirical efforts are the closest relatives of this work.
ProtocolBench~\citep{du2026protocolbench} benchmarks A2A, ACP, ANP, and
Agora on task success, end-to-end latency, message overhead, and
robustness across multi-agent workloads; MCP is explicitly out of scope,
and its timeline instrumentation (send, queue, service, first-token) does
not isolate connection establishment. \citet{dobrovolskyi2026empirical}
compares an MCP-based single-agent architecture against A2A multi-agent
architectures end-to-end over 450 task executions. That study attributes
part of the difference to setup costs, but it infers them rather than
instrumenting them. Neither work decomposes or measures the handshake
itself. Our benchmark is
complementary to both. Task-level benchmarks tell you what a protocol
costs once agents are talking, and ours tells you what it costs to start
talking.

The Agora protocol~\citep{marro2024agora} occupies a different corner of
the design space: agents negotiate protocols in natural language, trading
tokens for flexibility. We do not benchmark live LLM-mediated negotiation
here (Section~\ref{sec:limitations}), but Agora's own measurements of
negotiation cost motivate our token-cost metric.

The closest living analogue is service discovery in microservices, where
the same costs are paid at a different layer. Envoy's xDS protocol
streams endpoint and cluster configuration to sidecars so that a caller
never pays discovery on the request path~\citep{envoy2026xds}, and
HTTP/2 connection reuse and multiplexing exist precisely to keep
per-exchange setup off that path~\citep{thomson2022http2}. Both are
architectural answers to the question we measure: how do you stop paying
to find out who you are talking to? The methodological precedent is also
close. \citet{zhu2023meshinsight} decompose service-mesh sidecar
overhead into its constituent components and predict it from
configuration, which is the same decompositional move we make for the
handshake. The difference is what the two layers can assume. A service
mesh operates over a known, centrally configured fleet, so discovery
state can be pushed ahead of time; open agent ecosystems have no such
control plane, and the counterpart is a stranger whose capabilities are
discovered on first contact. That is why the costs we measure are paid
per pair rather than amortized fleet-wide.

Classical connection-establishment measurement is a direct methodological
ancestor. \citet{naylor2014cost} quantified the latency cost of the TLS
handshake at Internet scale, \citet{lee2021tls} measured TLS~1.3's
handshake-time reduction in deployment, and QUIC folded transport and
cryptographic setup into one round trip with 0-RTT
resumption~\citep{iyengar2021quic}. Pre-LLM multi-agent systems paid
setup costs too. Contract Net's announce--bid--award cycle is a
capability-discovery round trip~\citep{smith1980contractnet}, and
FIPA-ACL interaction protocols begin with capability registration and
service discovery~\citep{fipa2002acl}. What is genuinely new in the LLM
era is that handshake artifacts are often consumed by a language model.
An agent card or tool list may be placed into an LLM's context window, so
handshake content has a per-token \emph{inference} cost that classical
protocol analysis never had to account for. Byte counts and token counts
are correlated but priced very differently; we therefore measure both.

\section{Defining the Agent Handshake}
\label{sec:defs}

We define the \emph{handshake} as all communication from first contact
until the initiating agent can send its first substantive task message
with justified confidence that the counterpart can handle it. We
decompose it into four phases:
(a)~\emph{discovery}: locating the counterpart and retrieving its
self-describing metadata;
(b)~\emph{capability exchange}: learning what the counterpart can do, at
whatever granularity the protocol exposes;
(c)~\emph{auth/session}: establishing identity and any session state; and
(d)~\emph{readiness}: any residual signaling required before the first
task message is valid.
A phase may be empty for a given protocol, and a single message may serve
more than one phase; we attribute each wire exchange to the phase that
constitutes its primary function.

The handshake ends at the \emph{task-ready} state. Because ``ready'' is
protocol-specific, we define it operationally per protocol
(Table~\ref{tab:taskready}) against a fixed canonical capability: an
\texttt{echo} operation that returns its input verbatim. The initiator is
task-ready when it has positively confirmed, from protocol metadata, that
the counterpart offers \texttt{echo}. Using a trivial capability keeps
task execution cost negligible, so the handshake dominates. We also
execute one echo call after each timed handshake, outside the timed
region, to validate that the ready state is real.

\begin{table*}[t]
\centering
\footnotesize\setlength{\tabcolsep}{4pt}
\caption{Task-ready definitions and observed cold-start wire sequences.
Request and response sizes are HTTP body bytes from instrumented runs.}
\label{tab:taskready}
\begin{tabular}{@{}>{\raggedright\arraybackslash}p{0.16\textwidth}@{\hspace{1.4em}}>{\raggedright\arraybackslash}p{0.34\textwidth}@{\hspace{1.4em}}>{\raggedright\arraybackslash}p{0.40\textwidth}@{}}
\toprule
Protocol & Task-ready condition & Observed wire sequence (cold) \\
\midrule
\mcpl{} & \code{initialize} response received; \code{initialized}
notification sent; \code{tools/list} parsed; \texttt{echo} present &
POST \code{initialize} (163\,B$\rightarrow$379\,B); POST
\code{initialized} (54\,B, 202); POST \code{tools/list}
(68\,B$\rightarrow$399\,B); 3 round trips \\
\addlinespace
\mcpm{} & \code{server/discover} result adopted; \code{tools/list}
parsed; \texttt{echo} present & POST \code{server/discover}
(245\,B$\rightarrow$431\,B); POST \code{tools/list}
(240\,B$\rightarrow$522\,B); 2 round trips \\
\addlinespace
A2A & Agent Card fetched and parsed; skill \texttt{echo} present in card &
GET \pth{/.well-known/agent-card.json} ($\rightarrow$443\,B); 1
round trip \\
\addlinespace
ACP & Manifest list fetched and parsed; agent \code{echo-agent} present &
GET \texttt{/agents} ($\rightarrow$507\,B); 1 round trip \\
\bottomrule
\end{tabular}
\end{table*}

Phase attribution follows the protocols' semantics. MCP has no
in-protocol discovery; the endpoint URL is assumed known, which is itself
a design choice. Its legacy handshake therefore spends one exchange on
session establishment (\code{initialize}, which also negotiates the
protocol version and capability flags and yields the session identifier),
one on readiness (\code{initialized}), and one on capability exchange
(\code{tools/list}). Modern MCP's \code{server/discover} performs
version and capability negotiation without creating a session; we
attribute it, and \code{tools/list}, to capability exchange. A2A and
ACP fold discovery and capability exchange into a single metadata fetch,
with capability confirmation a local parse. Neither establishes a session
or requires auth in the anonymous configuration we benchmark.

We exclude from handshake measurement everything below the application
layer (TCP and TLS setup, reported separately as a baseline in
 Section~\ref{sec:results-baseline}), task execution itself, and
connection teardown. We do count the TCP connections each handshake
opens, since each one implies an additional transport round trip on a
real network.

\section{Methodology}
\label{sec:method}

\subsection{Harness}

The benchmark harness runs each protocol's official Python SDK client
against a minimal local counterpart agent built with the same project's
official server SDK. Counterparts are deliberately minimal, exposing
static capability metadata plus the echo capability, so that measured
costs are protocol floors rather than implementation artifacts. All three
SDKs use \texttt{httpx} for HTTP, which gives us a uniform
instrumentation point: a recording transport wraps every request with
monotonic timestamps, body sizes, and reconstructed wire sizes, and tags
it with a handshake phase. All traffic additionally traverses a local TCP
relay that counts exact bytes and connections per run.

Network conditions are emulated in the relay. For the \emph{50\,ms RTT}
condition it delays each forwarded chunk by 25\,ms per direction,
approximating a symmetric 50\,ms round-trip link at the application
layer. (We use a userspace relay rather than kernel-level emulation such
as \texttt{tc netem} for portability; Section~\ref{sec:validity}
discusses the difference.) The \emph{localhost} condition uses the same
relay with zero added delay, isolating protocol-stack overhead. For the
cold-start scenario we additionally sweep the emulated RTT across 0, 25,
50, and 100\,ms ($N{=}\SweepN{}$ runs per protocol per point, with fresh
counterpart servers per cell) to measure how latency-to-ready scales with
round-trip time (Figure~\ref{fig:f5}).

Each run is fully cold on the client side: a fresh SDK client object,
fresh HTTP connection pool, and fresh relay. Counterpart server processes
persist across the runs of a cell, as a long-running service would, and
five warmup runs precede each cell and are discarded. Timing uses a
monotonic nanosecond clock. The handshake timer starts immediately before
the client initiates first contact and stops when the task-ready
condition is confirmed. Every run is logged as one JSON record containing
the full request/response event list, and all statistics in this paper
are computed from those records by the released analysis scripts; the
numbers in the text are generated macros, not hand-typed values.

\subsection{Scenarios}

\textbf{S1: cold start.} Stranger agents. No prior contact, no cached
metadata, full discovery to task-ready. This is the headline number.

\textbf{S2: warm repeat.} The same pair, subsequent connection. Each
protocol is allowed exactly the reuse its specification and official SDK
support. For \mcpl{} that is nothing: response caching arrived only with
the modern era, so the SDK re-runs the full three-round-trip handshake,
tool listing included. For \mcpm{} we measure two policies. Under
\emph{auto}, the SDK's default connect policy, the client re-probes
\code{server/discover} and serves \code{tools/list} from the SEP-2549
response cache; under \emph{pinned}, it additionally remembers the
counterpart's protocol era and adopts it directly, reaching task-ready
entirely from cache. For A2A and ACP the client reuses its cached Agent
Card or manifest. Our MCP counterpart advertises a one-hour public cache
TTL, the configuration a latency-conscious deployment would choose; the
protocol default is non-cacheable.

\textbf{S3: capability mismatch.} The initiator requires a
\code{translate} capability the counterpart does not offer. We measure
time-to-rejection, that is, how much of the handshake must be paid before
the mismatch is discoverable.

\textbf{S4: degraded conditions.} (a)~\emph{Slow counterpart}: the
counterpart delays every response by 2\,s. (b)~\emph{Malformed metadata}:
the counterpart returns its capability metadata (initialize/discover
result, Agent Card, or manifest) truncated at 50\%, yielding
syntactically invalid JSON with valid HTTP framing. (c)~\emph{Version
mismatch}: the counterpart advertises only a fictitious protocol version
(2099-01-01 for MCP; 99.0.0 in the A2A Agent Card). ACP's manifest
carries no protocol-version field, so (c) is not testable for ACP, which
is itself a finding. For each degraded cell we record outcome, time to
failure or readiness, client retries, and the error surfaced to the
caller.

\subsection{Metrics and reporting}

For every scenario--protocol--network cell we report the following:
application-layer round trips to task-ready (request--response exchanges,
including fire-and-forget notifications); wall-clock latency to
task-ready, decomposed by phase; TCP connections opened; exact bytes on
the wire in each direction (relay-counted, through task-ready); and, for
degraded cells, retries and time-to-failure. We run $N{=}\NRuns$ measured
runs per cell and report medians and 95th percentiles throughout, never
bare means. Distributions were inspected for bimodality; none required
special treatment after warmup exclusion. Every median carries a
percentile bootstrap 95\% confidence interval over \CiIters{} resamples,
which separates two error sources that are easy to conflate. Within a
cell, sampling precision is high: the median CI half-width is
\CiMedHalfWidth{}\,ms and the largest is \CiMaxHalfWidth{}\,ms, so a
figure quoted to 0.1\,ms is resolved by the data. \emph{Across} cells,
client-side setup time drifts by a few milliseconds over a long session
(Section~\ref{sec:validity}), and that systematic term, not sampling
noise, is what limits between-protocol comparison. We therefore quote
medians at the precision the within-cell interval supports and state the
cross-cell floor explicitly wherever we compare cells
(Section~\ref{sec:results-s3}).

For token cost we ask what the handshake costs when an LLM-driven
initiator reads the counterpart's capability metadata, the normal case
when a model chooses tools or validates a counterpart. We take the exact
metadata bodies observed on the wire, count tokens with the open
\texttt{o200k\_base} tokenizer, and price them at a mid-tier frontier
model's list input rate (Claude Sonnet~5, \$3.00 per million input
tokens).\footnote{Anthropic pricing documentation,
\url{https://platform.claude.com/docs/en/about-claude/pricing}, accessed
2026-08-11. Introductory pricing of \$2 per million input tokens is in
effect through 2026-08-31; the standard \$3.00 rate takes effect
2026-09-01. We use the standard rate; all dollar figures scale linearly
with the assumed price.} Anthropic's own tokenizer counts roughly
15--20\% more tokens on JSON-like content than \texttt{o200k\_base}, so
the dollar figures are conservative lower bounds. Byte counts are exact.

\subsection{Environment and versions}

All experiments ran on a single Apple-silicon machine (macOS, Darwin
25.5.0, arm64) on 2026-08-10, with client, relay, and counterpart on
loopback. Python 3.12.13; \texttt{mcp} 2.0.0 (SDK released alongside spec
revision 2026-07-28); \texttt{a2a-sdk} 1.1.2 (A2A spec v1.0.0, JSON-RPC
binding); \texttt{acp-sdk} 1.0.3 (final release); \texttt{httpx} 0.28.1;
\texttt{uvicorn} 0.34.3; \texttt{fastapi} 0.115.14. Exact pins and the
one-command reproduction path are in the repository; the
appendix lists per-protocol launch details.

\subsection{Threats to validity}
\label{sec:validity}

\emph{Toy counterparts.} Our agents expose one capability. Real
counterparts with large tool inventories would shift byte and token costs
upward, though handshake latency is less affected, since inventory size
changes payloads rather than round trips. Our numbers are protocol
floors.
\par
\emph{Userspace network emulation.} The relay delays application-layer
chunks, not packets. TCP connection establishment against the relay
completes at loopback speed, so our 50\,ms-RTT figures exclude transport
setup entirely; on a real 50\,ms link each TCP connection would add one
additional RTT (plus one for TLS~1.3), which penalizes exactly the
protocols that open more connections (Section~\ref{sec:results-s1}).
SSE-framed responses cross the relay in two chunks and therefore pay the
per-chunk delay twice, which parallels, but does not perfectly model,
per-packet kernel emulation.
\par
\emph{Single SDK generation.} We measure specific SDK versions at a
specific date. SDK-added behaviors (connection handling, retries,
caching) are part of what we measure, deliberately, because they are what
deployed agents actually pay; results will drift as SDKs evolve.
\par
\emph{Single machine, loopback.} Absolute low-RTT latencies reflect one
machine's stack. The round-trip and byte counts, which drive the
higher-RTT results, are machine-independent. Client-side setup time also
drifts by a few milliseconds across cells within a long session, setting
a $\sim$5\,ms floor on cross-cell latency comparisons
(Section~\ref{sec:results-s3}).
\par
\emph{Token accounting.} Token costs model a hypothetical LLM-in-the-loop
reader with an open tokenizer at one vendor's list price. They scale
linearly in both assumptions and are stated as lower bounds.

\section{Results}
\label{sec:results}

\begin{table*}[t]
\centering
\footnotesize\setlength{\tabcolsep}{4pt}
\caption{Main results. Medians over $N{=}\NRuns$ runs per cell; RT =
application-layer round trips to task-ready (or to rejection for S3);
Conn = TCP connections opened; wire bytes are relay-counted through
task-ready (localhost condition shown). p95 in adjacent columns.}
\label{tab:main}
\begin{tabular}{llrrrrrrrr}
\toprule
 & & & & \multicolumn{2}{c}{localhost (ms)} & \multicolumn{2}{c}{50\,ms RTT (ms)} & \multicolumn{2}{c}{wire bytes} \\
\cmidrule(lr){5-6}\cmidrule(lr){7-8}\cmidrule(lr){9-10}
Scenario & Protocol & RT & Conn & med & p95 & med & p95 & up & down \\
\midrule
S1 cold start & MCP (legacy) & 3 & 3 & 6.1 & 8.0 & 230 & 236 & 1147 & 1522 \\
 & MCP (modern) & 2 & 1 & 4.4 & 5.2 & 122 & 125 & 1070 & 1205 \\
 & A2A & 1 & 1 & 2.8 & 3.0 & 61.0 & 61.9 & 170 & 569 \\
 & ACP & 1 & 1 & 2.8 & 3.2 & 62.1 & 63.1 & 149 & 633 \\
\addlinespace
S2 warm repeat & MCP (legacy) & 3 & 3 & 6.1 & 6.5 & 230 & 234 & 1147 & 1522 \\
 & MCP (modern, auto) & 1 & 1 & 3.6 & 4.2 & 63.5 & 65.4 & 540 & 557 \\
 & MCP (modern, pinned) & 0 & 0 & 2.1 & 2.6 & 6.6 & 7.3 & 0 & 0 \\
 & A2A & 0 & 0 & 2.0 & 2.7 & 2.0 & 2.7 & 0 & 0 \\
 & ACP & 0 & 0 & 2.1 & 2.5 & 6.6 & 7.4 & 0 & 0 \\
\addlinespace
S3 mismatch & MCP (legacy) & 3 & 3 & 6.2 & 6.9 & 231 & 235 & 1147 & 1522 \\
 & MCP (modern) & 2 & 1 & 4.4 & 5.5 & 117 & 123 & 1070 & 1205 \\
 & A2A & 1 & 1 & 2.9 & 4.3 & 62.2 & 65.3 & 170 & 569 \\
 & ACP & 1 & 1 & 2.9 & 4.0 & 57.7 & 63.2 & 149 & 633 \\
\bottomrule
\end{tabular}
\end{table*}

\subsection{S1: cold start}
\label{sec:results-s1}

\begin{figure*}[t]
\centering
\includegraphics[width=\textwidth]{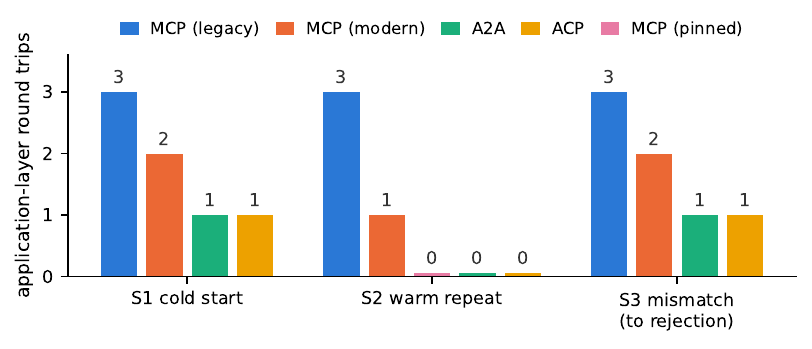}
\caption{Application-layer round trips to task-ready (S1, S2) and to
rejection (S3), by protocol variant. Warm values are what each protocol's
specification and official SDK allow a reconnecting client to skip; the
pinned MCP variant exists only for warm reconnects and so appears only in
the S2 group.}
\label{fig:f1}
\end{figure*}

Table~\ref{tab:main} and Figure~\ref{fig:f1} give the cold-start picture.
The protocols reach task-ready in three (\mcpl{}), two (\mcpm{}), and one
(A2A, ACP) application-layer round trips. On loopback, where a round trip
costs microseconds, the differences are visible but small in absolute
terms: medians of \SOneMcpLegacyLocalMed{}\,ms,
\SOneMcpModernLocalMed{}\,ms, \SOneAtoaLocalMed{}\,ms, and
\SOneAcpLocalMed{}\,ms respectively. Client-side protocol-stack work
dominates these numbers, not the wire.

Under the emulated 50\,ms RTT the ordering is unchanged but the spread
widens to exactly what the round-trip counts predict:
\SOneMcpLegacyRttMed{}\,ms for \mcpl{} (p95 \SOneMcpLegacyRttPNF{}\,ms),
\SOneMcpModernRttMed{}\,ms for \mcpm{}, \SOneAtoaRttMed{}\,ms for A2A,
and \SOneAcpRttMed{}\,ms for ACP. The legacy MCP handshake costs
\HeadlineColdRatio{}$\times$ A2A's. Two details deserve note. First,
\mcpl{} exceeds the naive $3 \times 50$\,ms prediction because two of its
three responses are SSE-framed and cross the relay in two delayed chunks;
measured against per-packet emulation the same effect appears as
serialization of header and body packets. Second, in the legacy era the
MCP client (\texttt{mcp} 2.0.0 defaults) opens a fresh TCP connection for
every JSON-RPC exchange: \SOneMcpLegacyLocalConns{} connections for the
legacy handshake versus one each for modern MCP, A2A, and ACP. We
verified this is client behavior rather than a harness artifact. The same
server reuses a single keep-alive connection both under raw sequential
HTTP requests and under the modern-era client, with or without the relay
in the path. The mechanism is visible in the SDK. Legacy-era responses
are SSE-framed, and the client closes each response stream as soon as the
JSON-RPC reply event arrives; an HTTP client cannot return an undrained
connection to its pool, so every exchange discards its connection. We
found no client-side configuration that restores reuse in the legacy era.
The modern era's plain-JSON responses restore it
automatically.\footnote{A related symptom of the same early-close pattern
was reported upstream as \texttt{python-sdk} issue \#2707 and closed for
lack of reproduction; we filed a deterministic reproduction
as issue \#3281,
\url{https://github.com/modelcontextprotocol/python-sdk/issues/3281}.}
Our relay hides the cost of the extra connections; on a real 50\,ms link
they would add roughly another 100\,ms of transport setup to the legacy
MCP handshake, and more if TLS is in play.

Wire volume tells a consistent but secondary story. Full cold handshakes
cost \SOneMcpLegacyLocalBytesUp{}+\SOneMcpLegacyLocalBytesDown{}\,B
up/down for \mcpl{} against
\SOneAtoaLocalBytesUp{}+\SOneAtoaLocalBytesDown{}\,B for A2A. That
difference matters at fleet scale but is invisible next to the round-trip
effect at interactive timescales.

\begin{figure*}[t]
\centering
\includegraphics[width=\textwidth]{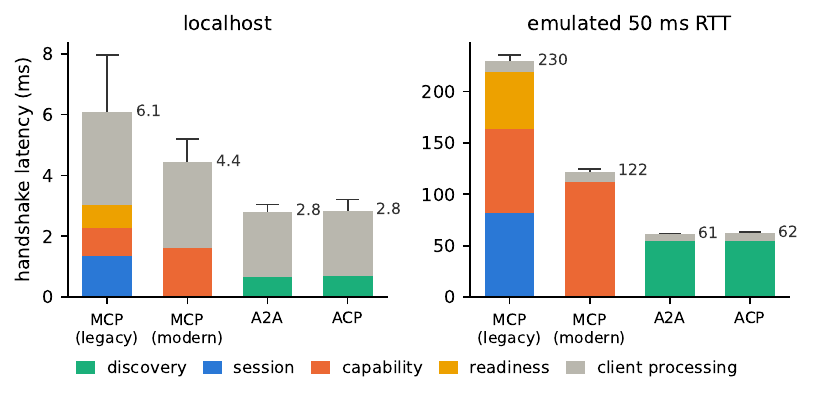}
\caption{S1 cold-start handshake latency decomposed by phase (median
stacked; whisker to p95 of the total), on loopback and under the emulated
50\,ms RTT. Note the differing y-scales: loopback cost is client
processing; at 50\,ms RTT the wire phases dominate in proportion to their
round trips.}
\label{fig:f2}
\end{figure*}

The phase decomposition (Figure~\ref{fig:f2}) locates the cost. On
loopback, most of every protocol's handshake is client processing: SDK
object construction, JSON parsing, validation. At 50\,ms RTT the picture
inverts. \mcpl{} spends its time in session establishment and capability
exchange (one round trip each, plus the readiness notification), \mcpm{}
in capability exchange (discover + list), and A2A/ACP in discovery, each
phase's share tracking its round trips. There is no hidden compute cost
that emerges at scale; the handshake is round trips nearly all the way
down.

\begin{figure}[t]
\centering
\includegraphics[width=\columnwidth]{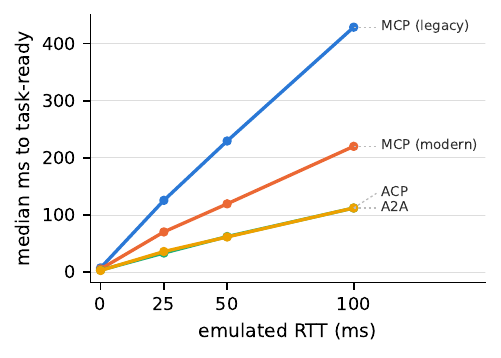}
\caption{Cold-start latency to task-ready versus emulated RTT (S1
medians, $N{=}\SweepN$ per cell). Slopes track the number of delayed
exchanges per handshake. Right-edge leader lines connect curves to labels
and are not data.}
\label{fig:f5}
\end{figure}

A sweep across emulated RTTs makes the mechanism direct
(Figure~\ref{fig:f5}). Median latency to task-ready rises linearly with
RTT, with fitted slopes of \SlopeAtoa{} (A2A), \SlopeAcp{} (ACP),
\SlopeMcpModern{} (\mcpm{}), and \SlopeMcpLegacy{} (\mcpl{}) milliseconds
of handshake per millisecond of RTT ($N{=}\SweepN$ per cell). The sweep's
50\,ms points, measured in separate cells from those of
Table~\ref{tab:main}, reproduce its medians within
\SweepVsTableMaxDeltaMs{}\,ms. The single-fetch protocols track their one
round trip almost exactly. The legacy MCP slope exceeds its three round
trips because its two SSE-framed responses (\code{initialize} and
\code{tools/list}) each cross the relay in two delayed chunks, while
the \code{initialized} notification returns a bare 202: three delayed
requests plus five delayed response chunks make eight one-way delays, or
four RTT equivalents, which is where the fitted \SlopeMcpLegacy{} lands
(Section~\ref{sec:validity}).

\subsection{S2: warm repeat and amortization}
\label{sec:results-s2}

Warm behavior separates the protocols more sharply than cold behavior
(Table~\ref{tab:main}, Figure~\ref{fig:f1}). \mcpl{} can skip nothing.
Its specification predates response caching, so a reconnecting client
repeats all \STwoMcpLegacyWarmRttRts{} round trips and pays
\STwoMcpLegacyWarmRttMed{}\,ms at 50\,ms RTT, within noise of its cold
handshake. \mcpm{} under the SDK's default policy re-probes
\code{server/discover} but serves the tool list from the SEP-2549
cache: \STwoMcpModernWarmRttRts{} round trip,
\STwoMcpModernWarmRttMed{}\,ms. A client that additionally pins the
counterpart's protocol era reaches task-ready with zero wire exchanges,
in \STwoMcpPinnedWarmRttMed{}\,ms. That is the same zero-round-trip
regime as A2A's cached-card (\STwoAtoaWarmRttMed{}\,ms) and ACP's
cached-manifest (\STwoAcpWarmRttMed{}\,ms) reconnects; the residual
handful of milliseconds in each case is client-side setup, not wire
exchange.

\begin{figure}[t]
\centering
\includegraphics[width=\columnwidth]{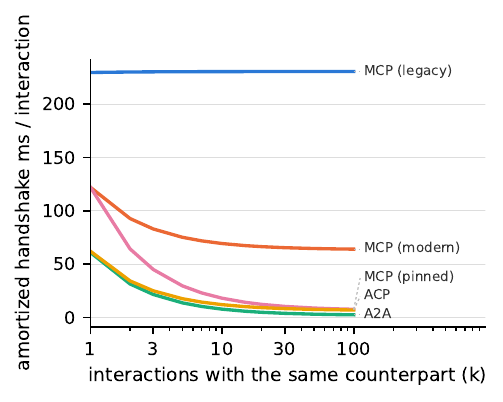}
\caption{Amortized handshake cost per interaction over $k$
interactions with the same counterpart (cold $+ (k{-}1)\times$ warm, over
$k$), under the emulated 50\,ms RTT. Protocols with zero-round-trip warm
paths amortize to a few milliseconds of client-side setup; \mcpl{} pays
its full handshake forever. Right-edge leader lines connect curves to
labels and are not data.}
\label{fig:f4}
\end{figure}

Figure~\ref{fig:f4} shows what this means over repeated interactions. At
$k{=}100$ interactions with the same counterpart under 50\,ms RTT, the
per-interaction handshake cost amortizes to \AmortMcpLegacy{}\,ms for
\mcpl{}, which is effectively no amortization at all, versus
\AmortMcpModern{}\,ms for \mcpm{} (auto), and \AmortMcpPinned{}\,ms,
\AmortAtoa{}\,ms, and \AmortAcp{}\,ms for pinned MCP, A2A, and ACP. For
agent pairs that interact repeatedly, cacheability, not cold-start cost,
is the design property that matters. For true strangers, the cold numbers
stand.

\subsection{S3: capability mismatch}
\label{sec:results-s3}

Time-to-rejection mirrors time-to-ready, because in all three protocols
the initiator discovers what the counterpart \emph{cannot} do the same
way it discovers what it can. A2A and ACP reject after their single
metadata fetch (\SThreeAtoaRttMed{}\,ms and \SThreeAcpRttMed{}\,ms at
50\,ms RTT), and \mcpm{} pays its discover-plus-list
(\SThreeMcpModernRttMed{}\,ms). \mcpl{} must complete its entire
three-round-trip handshake, session included, before learning the
counterpart is useless to it (\SThreeMcpLegacyRttMed{}\,ms). The wasted
cost of a mismatched contact is thus the full handshake price, and it
compounds for agents that shop across many counterparts.
Document-centric discovery makes ruling out a counterpart as cheap as one
GET; session-first designs charge the session price merely to browse. The
ACP medians invert slightly across these two scenarios
(\SThreeAcpRttMed{}\,ms to rejection versus \SOneAcpRttMed{}\,ms to cold
task-ready). The per-request logs attribute the entire difference to
client-side setup before the request leaves the client (wire-level GET
times agree to within a millisecond), which drifts by a few milliseconds
between cells over a long benchmarking session. Cross-cell differences
below roughly 5\,ms should be read as this noise floor, not as protocol
effects. Note that this is a systematic cross-cell term, not sampling
error: the bootstrap intervals within each cell are an order of magnitude
tighter (Section~\ref{sec:method}), which is why the two cells can be
individually well resolved and still not differ meaningfully from each
other.

\subsection{S4: degraded conditions}

\begin{table*}[t]
\centering
\footnotesize\setlength{\tabcolsep}{4pt}
\caption{Behavior under degraded conditions ($N{=}\NRuns$ per cell,
localhost). S4a reports time to task-ready with a 2\,s per-response
server delay; S4b/S4c report time to failure. Retr.\ = client retries
observed (duplicate requests). For S4c, ACP is untestable: its manifest
carries no protocol-version field.}
\label{tab:degraded}
\begin{tabular}{llllrl}
\toprule
Scenario & Protocol & Outcome & med.\ time (ms) & Retr. & Error surfaced \\
\midrule
S4a slow (2\,s/resp.) & MCP (legacy) & ready & 6026 & 0 & none \\
 & MCP (modern) & ready & 4019 & 0 & none \\
 & A2A & ready & 2011 & 0 & none \\
 & ACP & ready & 2011 & 0 & none \\
\addlinespace
S4b truncated metadata & MCP (legacy) & error & 3.5 & 0 & \texttt{MCPError} \\
 & MCP (modern) & error & 4.4 & 0 & \texttt{MCPError} \\
 & A2A & error & 3.0 & 0 & \texttt{AgentCardResolutionError} \\
 & ACP & error & 2.9 & 0 & \texttt{JSONDecodeError} \\
\addlinespace
S4c version mismatch & MCP (legacy) & error & 3.7 & 0 & \texttt{RuntimeError} \\
 & MCP (modern) & error & 4.2 & 0 & \texttt{RuntimeError} \\
 & A2A & ready & 3.0 & 0 & none (version deferred) \\
\bottomrule
\end{tabular}
\end{table*}

Table~\ref{tab:degraded} summarizes the degraded scenarios.

\textbf{S4a: slow counterpart.} With 2\,s injected per response, time to
task-ready is a clean multiple of round-trip count:
\SFourAMcpLegacyLocalMed{}\,ms for \mcpl{} (three delayed responses),
\SFourAMcpModernLocalMed{}\,ms for \mcpm{}, and about 2\,s for A2A and
ACP (\SFourAAtoaLocalMed{}\,ms, \SFourAAcpLocalMed{}\,ms). No SDK timed
out or retried; the delay simply multiplies. A slow counterpart is
\emph{most} expensive to the protocol with the most round trips, and the
initiator has no way to discover the slowness before paying the full
handshake.

\textbf{S4b: malformed metadata.} All four variants fail fast, with no
retry storms and no hangs. Every client detected truncated JSON on first
parse, in milliseconds. What differs is error quality. A2A's SDK raises a
typed \code{AgentCardResolutionError} naming the URL. MCP's raises a
protocol-level \code{MCPError} wrapping the JSON validation failure
(the modern client, having received a malformed discover response, falls
back to \code{initialize} per its negotiation rule and fails cleanly
when that response is also malformed: two requests, no blind retries).
ACP's client surfaces a bare \code{JSONDecodeError} with a character
offset and no indication of which endpoint or agent was involved. All
three abort rather than degrade, which is the right behavior; only ACP
leaves the caller to reconstruct the context.

\textbf{S4c: version mismatch.} Here the protocols genuinely diverge.
Both MCP eras refuse a counterpart that speaks only the fictitious
version 2099-01-01, in
\SFourCMcpLegacyLocalTtfMed{}--\SFourCMcpModernLocalTtfMed{}\,ms, with an
explicit error naming the unsupported version; the modern probe's
structured \texttt{-32022} error even carries the server's
supported-version list. A2A's client behaved differently. It fetched a
card advertising protocol version 99.0.0, selected the JSON-RPC binding,
and completed echo tasks in normal handshake time
(\SFourCAtoaLocalMed{}\,ms) without surfacing any warning. This is
spec-conformant client behavior, not an SDK bug. The normative schema
marks the interface's \code{protocolVersion} as descriptive (``the
version of the A2A protocol this interface exposes''), and A2A's
mandatory version negotiation instead happens per request, via an
\code{A2A-Version} header that servers must reject when
unsupported~\citep{a2aspec2026}. On the wire we observed exactly that
split. The client's task request carried \texttt{a2a-version: 1.0}, its
own version, independent of the card's claim, and our genuinely v1.0
counterpart accepted it. The consequence for handshake design stands:
discovery-time version claims go unvalidated, so a real incompatibility
surfaces not during the handshake but at the first task message, as a
server-side rejection, and when the wire versions happen to agree, as
here, it never surfaces at all. ACP has no version signal to check.
Deferred version checking is convenient exactly until an incompatibility
is real, at which point what could have been a cheap, legible handshake
failure becomes a task-time failure attributed to the wrong layer.

\subsection{Context cost of capability metadata}
\label{sec:results-tokens}

\begin{table}[t]
\centering
\small
\caption{Capability metadata an LLM-driven initiator ingests during one
cold handshake with our one-capability counterpart (exact wire bodies;
\texttt{o200k\_base} tokens, a lower bound; see
Section~\ref{sec:method}).}
\label{tab:tokens}
\begin{tabular}{lrrr}
\toprule
Protocol & metadata bytes & tokens & USD / $10^6$ handshakes \\
\midrule
MCP (legacy) & 726 & 181 & \$543 \\
MCP (modern) & 953 & 242 & \$726 \\
A2A & 443 & 113 & \$339 \\
ACP & 507 & 112 & \$336 \\
\bottomrule
\end{tabular}
\end{table}

\begin{table}[t]
\centering
\small
\caption{Handshake metadata in tokens as the counterpart's capability
inventory grows, and the marginal cost of one capability. Built by
replicating the captured wire entry $n$ times inside the byte-identical
envelope, so the documents are exactly what a counterpart with $n$ such
capabilities would serve.}
\label{tab:inventory}
\begin{tabular}{lrrrrr}
\toprule
Protocol & $n{=}1$ & $n{=}10$ & $n{=}50$ & $n{=}200$ & per cap. \\
\midrule
MCP & 96 & 798 & 3,918 & 15,618 & 78 \\
A2A & 115 & 331 & 1,291 & 4,891 & 24 \\
ACP & 112 & 1,075 & 5,355 & 21,405 & 107 \\
\bottomrule
\end{tabular}
\end{table}

When the initiating agent is an LLM, the counterpart's metadata is not
merely transferred, it is \emph{read}: placed in a context window to
select a tool, check a skill, or validate a counterpart. The scarce
resource is therefore context, and the relevant question is what fraction
of the window the handshake consumes before any task content arrives.

For our one-capability counterpart the answer is trivial
(Table~\ref{tab:tokens}): \TokAcp{}--\TokMcpModern{} tokens, a rounding
error in any modern context window. That number is a protocol floor and
should not be quoted as a cost. It matters only as an intercept.

The slope is the finding (Table~\ref{tab:inventory}). Each additional
capability costs \InvPerCapAtoa{} tokens in an A2A card,
\InvPerCapMcp{} in an MCP tool listing, and \InvPerCapAcp{} in an ACP
manifest. At \InvMaxN{} capabilities, a plausible size for a mature tool
server, the handshake alone delivers \InvMcpNTwoHundred{} tokens for MCP and
\InvAcpNTwoHundred{} for ACP, which is \InvMaxCtxPct{}\% of a 200k-token window
spent before the agent has read its own task. Three points follow.

First, the per-capability costs differ by more than a factor of four, and
the ordering is the reverse of the one-capability ranking. A2A is
cheapest per skill because a card entry is a name, a description, and
tags; MCP is dearer because a tool entry carries full input and output
JSON schemas; ACP is dearest because each agent entry repeats a
sixteen-field metadata block that is null in our counterpart and would be
populated in a real one. A protocol's floor tells you little about what it
costs at deployment scale.

Second, this is the one handshake cost that round-trip reduction does not
touch. It is carried by the \emph{listing} step, which every protocol
performs and which the 2026 MCP revision made cacheable but no cheaper.
Indeed the modern era spends tokens to save round trips: its richer
discovery envelope makes it the most token-expensive handshake at $n{=}1$
(\TokMcpModern{} versus \TokMcpLegacy{}), a trade that pays for itself
only because the result is cacheable.

Third, cacheability is what bounds the damage. The same
SEP-2549/cached-card machinery that eliminates warm round trips keeps
already-ingested metadata valid, so an agent that reconnects pays the
context cost once rather than per interaction. For an agent that shops
across many counterparts, however, there is no reuse to be had, and
inventory size becomes the binding constraint on how many counterparts it
can consider at all.

\subsection{Transport-layer baseline}
\label{sec:results-baseline}

For calibration: on the same loopback, a bare TCP connect completes in
\BaselineTcpMed{}\,ms (median) and a full TLS~1.3 handshake in
\BaselineTlsMed{}\,ms of compute. On a real network each adds one round
trip~\citep{rescorla2018tls}. Against that two-round-trip
transport-plus-cryptographic stack, A2A and ACP add a single
application-layer round trip, half the cost of the entire stack beneath
them. A legacy MCP handshake adds three, exceeding the combined transport
and cryptographic setup it rides on, and that is before any
authentication, which none of our configurations exercised. The
application layer has, in effect, re-created the pre-TLS-1.3 handshake
stack one level up, and it is now re-learning the same remedies (fewer
round trips, resumable state) that transport designers adopted a decade
ago~\citep{lee2021tls,iyengar2021quic}.

\section{Discussion}
\label{sec:discussion}

\textbf{Where the overhead concentrates.} Our measurements are consistent
across scenarios. Handshake cost is round trips first, connections
second, bytes a distant third, and compute negligible. Every design
question about agent handshakes should therefore be asked as ``how many
round trips does this add, and can it be cached?'' The protocols' own
trajectories agree. MCP's 2026 revision is precisely a
round-trip-reduction release: it replaced a three-message, session-first
initialization with a one-probe, cacheable discovery, and our
measurements quantify the effect at \McpModernSpeedup{}\% off the cold
handshake at 50\,ms RTT and, with pinning, free warm reconnects. That the
youngest protocol in our study already shipped this change is itself
evidence that handshake overhead is a live design pressure, not a
hypothetical.

\textbf{Document-centric versus session-first discovery.} A2A and ACP
demonstrate the value of putting the entire capability surface in one
cacheable document at a well-known location: cold task-ready in one round
trip, mismatch rejection at the same price, warm reconnects free. This
design has two costs. One is currency, since a cached card can be stale
and neither protocol gives the initiator a cheap way to validate
freshness short of re-fetching. The other is expressiveness, since a
static document cannot answer negotiated or per-principal capability
questions. MCP's modern era lands in between. It keeps a negotiated,
session-capable protocol but makes the negotiation's \emph{results}
cacheable with explicit server-controlled TTLs, which preserves freshness
semantics the static documents lack.

\textbf{Failure legibility is a protocol property.} The degraded
scenarios show that what a protocol checks is as consequential as what it
exchanges. MCP's explicit version negotiation bought it clean, fast,
well-attributed failures in every degraded cell. A2A, by design, defers
version enforcement to per-request headers checked by the server, so its
discovery-time version claims pass unexamined. ACP's missing version
field forecloses the check entirely. For autonomous agents, which cannot
eyeball a stack trace, the difference between a structured ``unsupported
version, here is what I support'' and a counterpart whose incompatibility
surfaces only at task time is the difference between a recoverable
negotiation and a debugging session.

\textbf{Implications for agent-system designers.} Three follow directly
from the data. First, treat counterpart metadata as a cacheable asset
with a lifecycle: the measured gap between \mcpl{} and everything else is
mostly the gap between re-deriving and reusing handshake state. Second,
budget for round trips at discovery time when agents shop across many
counterparts. S3 shows rejection costs a full handshake, so a design that
can rule counterparts out from a one-GET document scales shopping
linearly in counterparts, while session-first designs scale it in
handshakes. Third, keep capability inventories lean if an LLM will read
them. Section~\ref{sec:results-tokens} shows the per-capability slope,
not the floor, is what binds: listing is the step that grows with
inventory, and it is the one handshake cost that caching defers but
round-trip reduction never touches. We deliberately stop
at implications. Whether a future protocol should, for example, merge
A2A-style cards with MCP-style negotiated sessions is a design question
our measurements inform but do not settle.

\section{Limitations and Future Work}
\label{sec:limitations}

Beyond the validity threats of Section~\ref{sec:validity}, three
boundaries mark natural extensions. We measured anonymous configurations;
authenticated handshakes (OAuth flows in MCP, credentialed A2A cards) add
round trips and token exchanges that deserve the same decomposition. We
measured one counterpart per protocol; sweeping capability-inventory size
would turn our token floors into curves. And we excluded live
natural-language negotiation in the style of
Agora~\citep{marro2024agora}, where two LLM-backed agents converge on a
shared task format before first use. By Agora's own accounting such
negotiation costs thousands of tokens per pair, orders of magnitude above
every structured handshake we measured, and benchmarking it under our
phase decomposition, including its amortization across subsequent
structured interactions, is the most interesting open cell in our matrix.
Finally, our S2 results suggest that hybrid designs (cacheable discovery
documents carrying negotiated session hints, or 0-RTT-style resumption
tokens for agent sessions) could combine the strengths we measured
separately. Evaluating such designs is future work, not a proposal we
make here.

\section{Conclusion}
\label{sec:conclusion}

To our knowledge, this is the first empirical measurement of the agent
handshake decomposed into its phases (discovery, capability exchange,
session, readiness), across the major AI agent communication protocols
and across cold, warm, mismatch, and degraded scenarios, with an open
harness and raw data. The tax of saying hello is real and structured. It is paid in round
trips, multiplied by network latency and counterpart slowness, waived by
cacheability, and, in one protocol's own 2026 revision, already being
engineered down. Agent-to-agent interaction is young enough that its
connection economics are still being designed. We hope a shared,
reproducible way to measure the cost of hello helps that design
conversation stay empirical.

\bibliographystyle{plainnat}

\bibliography{refs}

\appendix

\section{Reproducibility}
\label{app:repro}

The repository at

\begin{center}
\url{https://github.com/pvarshh/handshake-benchmark}
\end{center}

\noindent
contains the harness (\texttt{/harness}), the minimal counterpart agents
(\texttt{/agents}), raw per-run JSONL for every cell
(\texttt{/results/raw}), and the scripts that generate every figure,
table, and in-text number (\texttt{/analysis}). Three commands
reproduce the raw data, and one regenerates every figure, table, and
number in this paper from it:

\begin{quote}
\small
\texttt{python -m harness.run\\
\hspace*{1em}-{}-scenario all -{}-n 50 -{}-network both}\\
\texttt{python -m harness.baseline}\\
\texttt{python -m harness.rtt\_sweep}\\
\texttt{python -m analysis.run\_all}
\end{quote}

\noindent
The first takes roughly 25 minutes on a laptop; the others are quick.
Version pins: Python 3.12.13,
\texttt{mcp} 2.0.0, \texttt{a2a-sdk} 1.1.2, \texttt{acp-sdk} 1.0.3,
\texttt{httpx} 0.28.1, \texttt{uvicorn} 0.34.3, \texttt{fastapi}
0.115.14, \texttt{starlette} 0.46.2. The MCP counterpart is served over
streamable HTTP with SEP-2549 cache hints (\code{ttlMs} =
3{,}600{,}000, i.e.\ one hour, public scope) on \code{server/discover}
and \code{tools/list}; the A2A counterpart serves a v1.0 Agent Card
with one skill over the JSON-RPC binding; the ACP counterpart serves one
agent manifest. Degraded variants are driven by environment flags
documented in the repository README.

\end{document}